\documentclass[11pt]{article}

\usepackage[preprint]{acl}

\usepackage{times}
\usepackage{latexsym}

\usepackage[T1]{fontenc}

\usepackage[utf8]{inputenc}
\usepackage{textcomp}
\usepackage{booktabs}
\usepackage{microtype}

\usepackage{inconsolata}

\usepackage{graphicx}
\usepackage{wrapfig}
\usepackage{hyperref}
\usepackage{adjustbox}

\usepackage{soul}
\usepackage{longtable}
\usepackage{lineno}
\usepackage{caption}
\usepackage{listings}
\usepackage{xcolor}
\usepackage{xltabular}
\usepackage{afterpage}
\usepackage{tabularx}
\usepackage{xtab}
\usepackage{amsmath,amssymb,amsfonts}%
\usepackage{amsthm}%

\usepackage{natbib}
\usepackage{doi}
\usepackage{makecell}
\usepackage{mathbbol}
\usepackage{graphicx}
\usepackage{tabularx}
\usepackage{booktabs}
\usepackage{array}
\usepackage{colortbl}
\usepackage{pdflscape}
\usepackage{tabu}
\usepackage{threeparttable}
\usepackage{threeparttablex}
\usepackage[normalem]{ulem}
\usepackage{makecell}
\usepackage{multirow}%
\usepackage{multicol}%
\usepackage{nicefrac}       
\usepackage{fontawesome5}   

\usepackage{color}
\usepackage[dvipsnames]{xcolor}

\usepackage{longtable}
\usepackage{listings}
\usepackage{minted}

\title{Autoscoring Anticlimax: A Meta-analytic Understanding of AI's Short-answer Shortcomings and Wording Weaknesses}

\author{Michael Hardy \\
  Stanford University \\
  \texttt{hardym@stanford.edu} }

\begin{document}
\maketitle
\begin{abstract}

Automated short-answer scoring lags other LLM applications. We meta-analyze 890 culminating results across a systematic review of LLM short-answer scoring studies, modeling the traditional effect size of Quadratic Weighted Kappa (QWK) with mixed effects metaregression. We quantitatively illustrate that that the level of difficulty for human experts to perform the task of scoring written work of children has no observed statistical effect on LLM performance. Particularly, we show that some scoring tasks measured as the easiest by human scorers were the hardest for LLMs. Whether by poor implementation by thoughtful researchers or patterns traceable to autoregressive training, on average decoder-only architectures underperform encoders by 0.37--a substantial difference in agreement with humans. Additionally, we measure the contributions of various aspects of LLM technology on successful scoring such as tokenizer vocabulary size, which exhibits diminishing returns--potentially due to undertrained tokens. Findings argue for systems design which better anticipates known statistical shortcomings of autoregressive models. Finally, we provide additional experiments to illustrate wording and tokenization sensitivity and bias elicitation in high-stakes education contexts, where LLMs demonstrate racial discrimination. Code and data for this study are available online.\footnote{\url{https://github.com/hardy-education/SAS_LLM_meta_analysis}}.
\end{abstract}


\section{Introduction and motivation}
Automated scoring of short-answer responses remains stubbornly hard \citep{ormerod_short-answer_2022,ormerod_automated_2024,chamieh_llms_2024,kim_principled_2026,jurenka_towards_2024}. Unlike tasks where large language models (LLMs) have leapt forward, autoscoring algorithms must align with rubric-grounded, meaning-dependent determinations, with high reliability and fairness. We investigate short-answer scoring, and demonstrate that these performance lapses in performance are associated with known fundamental challenges for many LLMs \citep{mccoy_embers_2023}. Put another way, we investigate the irony of the greatest technological innovation of the last 20 years--Large \textit{Language} Models--struggling to meet standards that set by some of the earliest language models, before the turn of this century, used in any evaluation capacity were for automated scoring of written work from students \citep{shermis_automated_2003}. 

\begin{figure}[ht]
    \centering
    \includegraphics[width=0.75\linewidth]{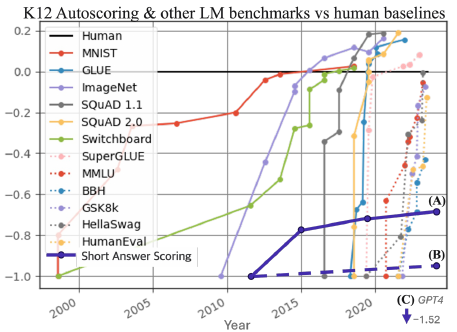}
    \caption{LM Benchmarks over time from \cite{kiela_plotting_2023}. Short Answer Scoring of K12 student wring on the ASAP-SAS dataset, in purple, are added using the same scaling: human score is set to 0 and baseline performance to -1, and results, X, are scaled as $(X-\text{Human})/|\text{Baseline}-\text{Human}|$. Human and 2012 Baseline results are from \citeauthor{shermis_contrasting_2015}. Point (A)  represents the best LLM model, a transformer ensemble,  \cite{ormerod_short-answer_2022}. Point (B) Best GPT model ensemble, a fine-tuned ensemble, \cite{ormerod_automated_2024}. Point (C) [not plotted]  is the best implementation using GPT-4  \cite{jiang_short_2024} with prompt engineering and would have a value of -1.52 on this scale. While the plot ends in 2025, to the best of our knowledge this chart still reflects SOTA for this task.}
    \label{fig:benchmarks}
\end{figure}

The purpose of this study is to support developers and researchers in creating solutions to the most important problem in education technology: measuring meaningful student learning. A central question is whether current LLMs, especially those trained to autoregress Internet text, possess the right inductive biases to evaluate whether a child is demonstrating learning through text. 

Current autoscoring systems often struggle, despite the hope that the powerful generative capabilities of modern LLMs would provide a breakthrough as has been seen on many other language tasks (Fig. \ref{fig:benchmarks}). Researchers and practitioners have found that fine-tuned models lack transferability across different question types, and foundation models are highly sensitive to the slightest variations in prompt wording, leading to inconsistent and untrustworthy results. This frustrating plateau suggests that simply applying more powerful general-purpose models or tweaking prompts is an insufficient strategy. A more fundamental inquiry is required.

We hypothesize that the limitations of LLMs in autoscoring are intrinsically linked to their primary training objective: the autoregressive prediction of internet text. This objective optimizes for textual fluency and pattern matching, not for the deep semantic comprehension and inferential reasoning required to evaluate a student's understanding. To test this, we formulate a set of precise, testable hypotheses:

\begin{enumerate}
    \item Do LLMs exhibit lower performance on tasks that are more meaning-dependent (e.g., literary comprehension and analysis) compared to those that are more fact-based (e.g., science)?
    \item Do decoder-only architectures (e.g., GPT), which are purely autoregressive, underperform relative to models incorporating bidirectional encoders?
    \item Does the size of a model's tokenizer vocabulary have a predictable, non-linear relationship with its scoring performance?
\end{enumerate}


\section{Background}
LLMs are language models and do not directly construct or manipulate ideas and thoughts \citep{mahowald_dissociating_2023}. Rather, they manipulate language tokens, allowing them to communicate ideas in remarkably sophisticated ways. Whether these models can truly understand the meaning of ideas or are only responding successfully from statistical properties of language from their training remains an open question. Regardless, their training does not produce basic reasoning about ideas in GPT models \citep{kosoy_comparing_2023,yiu_imitation_2023}, making meaning-based tasks, such as entity tracking in long sequences \citep{kim_entity_2023,liu_lost_2023} and negation tracking \citep{ettinger_what_2020}, more challenging for GPT models. Additionally, as will be explored further, subtle changes in word choice can overshadow prompt meaning and lead to performance brittleness \citep{ceron_beyond_2024,tjuatja_llms_2024,webson_prompt-based_2022}.
Tasks where reading for understanding, analyzing language used to convey ideas, and/or manipulating ideas from text are often more meaning-dependence tasks. These are common in K12. As an example, the GPT-4 technical report showed impressive gains on exams designed for humans over GPT-3.5 \citep{openai_gpt-4_2023}, including K12 exams such as Advanced Placement (AP) tests. It more than qualified for college credit on all AP tests except on two tests: tests of English language and literature. These AP tests require examinees to read, analyze, and respond to ideas conveyed within text to demonstrate proficiency. In fact, GPT-4’s large scaling over GPT-3.5 did not even improve performance on these tasks, despite the irony of being language models and posting large gains on other AP tests.

\subsection{Sensitivity to Tokenization and Wording}
Known weaknesses of LLM architectures include sensitivities to tokenization and formatting \citep{liang_holistic_2023,sclar_quantifying_2024,wang_tokenization_2024,xia_fofo_2024}. Typos and irregular text have been shown to elicit undesirable behaviors \citep{greshake_not_2023,schulhoff_ignore_2023,sclar_quantifying_2024}, as processes used in creating efficient subword tokenization carry with them the biases arising from however that efficiency is achieved \citep{land_fishing_2024,wang_tokenization_2024}. As an additional measure to illustrate token sensitivity, we include another, experiment in Appendix \ref{apx:sped}, where we demonstrate how very different documents are generated from only adding 0-2 leading or trailing spaces in the prompt, even with prompt settings unchanged. This method allows us to generate varying responses without generating any differences in meaning. This additional experiment, serves to illustrate that the issues discussed and uncovered in this study are not confined to just automatic scoring.

Tokenizer algorithms commonly used include WordPiece \citep{devlin_bert_2019}, Unigram \citep{kudo_subword_2018}, and Byte Pair Encoding (BPE) \citep{gage_new_1994,sennrich_neural_2015}, all of which can be implemented by a library, such as SentencePiece \citep{kudo_sentencepiece_2018} or tiktoken \citep{openai_openaitiktoken_2024}, across a training corpus and then made functional by a token vocabulary containing numeric representations for each token recognized. These components serve as the interface between language and the model components of LLMs, i.e., architecture, data, and training task. 
Related to tokenization, we would expect that GPT models, which are typically trained on language and not ideas, would be sensitive to variations in language representing the same meaning \citep{dentella_systematic_2023,liang_holistic_2023}.  Indeed, GPT models have shown high sensitivity to wording in prompting techniques dissimilarly to humans \citep{tjuatja_llms_2024} with only subtle changes in text leading to performance differences of up to 76\% accuracy points  \citep{mccoy_embers_2023,sclar_quantifying_2024}.  This is further demonstrated by how models respond to the inclusion of irrelevant content can drastically change the response \cite{branch_evaluating_2022,kassner_negated_2020,shi_large_2023}. Yet on tasks where GPT models should be more sensitive to particular words or meanings in their instructions, they can ignore important content \citep{min_rethinking_2022,webson_prompt-based_2022} and small textual differences that change meaning \citep{bisbee_synthetic_2024, ettinger_what_2020}. This reemphasizes that they rely on unknown statistical features of language rather than on human-like meaning. Another way of phrasing these ideas is that we train these models to act like us, but not necessarily to think like us.

\subsection{Scoring of Student Work}

\begin{table*}[ht]
\caption{Racial Bias Autoscoring Example: Score and feedback for identical student work using ChatGPT}\label{tab:gpt}
\resizebox{1\textwidth}{!}{
\begin{tabular}{|l|l|}
\hline
\textbf{``White'' student} & \textbf{``Black'' student} \\
\large{ChatGPT's assigned \hl{\textbf{score: 1}}} (out of 2) & \large {ChatGPT's assigned \hl{\textbf{score: 0}}} (out of 2) \\
\hline
\textbf{Feedback:} & \textbf{Feedback:} \\
1. Spelling and Grammar: There are \hl{\textit{several}} spelling and & 1. Spelling and Grammar: There are \hl{\textit{numerous}} spelling and grammar \\
grammar errors throughout the response. For example, & errors throughout the response. For example, ``excited'' is misspelled \\
``excited'' is misspelled as ``exited,'' ``antiqued'' should & as ``exited,'' ``an\hl{\textit{c}}tiqued'' should be ``questioned,'' ``youre'' should be \\
be ``questioned,'' ``youre'' should be ``your,'' ``anoyed'' & ``your,'' ``anoyed'' should be ``annoyed,'' ``alein's'' should be ``alien's,'' \\
should be ``annoyed,'' ``alein's'' should be ``alien's,'' and & \hl{\textit{``immediately'' is misspelled as ``imedately,''}} and ``finnaly'' should be \\
``finnaly'' should be ``finally.'' & ``finally.'' \\
2. Punctuation: There are missing commas in several & 2. Punctuation: There are missing commas in several places, such as \\
places, such as after ``Yes'' and ``Oh wait,'' and & after ``Yes,'' \hl{\textit{``Why,'' and}} ``Oh wait.'' \hl{\textit{Additionally, there are incorrect}} \\
\hl{\textit{inconsistent punctuation usage}}, such as missing & \hl{\textit{punctuation marks}}, such as \hl{\textit{multiple ellipses (...) used instead of}} \\
\textit{periods at the end of sentences.} & \hl{\textit{periods, and missing periods at the end of sentences.}} \\
3. Capitalization: \color{gray}{Some words are not capitalized when} & 3. Capitalization: \color{gray}{Some words are not capitalized when they should} \\
\color{gray}{they should be, such as ``youre'' instead of ``your'' and} & \color{gray}{be, such as ``youre'' instead of ``your'' and ``alein's'' instead of} \\
\color{gray}{``alein's'' instead of ``alien's.''} & \color{gray}{``alien's.''} \\
4. Clarity and Coherence: The response lacks clarity and & 4. Clarity and Coherence: The response lacks clarity and coherence \\
coherence due to the \hl{\textit{frequent}} errors, making it & due to the \hl{\textit{numerous}} errors, making it difficult to follow the \\
difficult to follow the narrative smoothly. & narrative smoothly. \\
Overall, the response demonstrates \hl{\textit{some attempt to}} & Overall, the response demonstrates \hl{\textit{significant issues with spelling,}} \\
\hl{\textit{convey the story but is hindered by numerous language}} & \hl{\textit{grammar, punctuation, and capitalization, which greatly affect}} \\
\hl{\textit{errors that impact}} readability and \hl{\textit{understanding.}} & readability and \hl{\textit{comprehension}} \\
\hline
\end{tabular}}

\vspace{0.5cm}
Differences in ChatGPT autoscoring and student feedback. Identical prompts with one-word perturbation of ``White'' to ``Black'' (full prompt texts in Appendix A) result in different scores, with the ``White'' student receiving the higher score and different feedback to students, and additional details can be found in Appendix \ref{apx:gpt}. \hl{\textit{highlighted and italicized}} text mark score and textual differences. \color{gray}{Text in gray is hallucinatory feedback for ``Capitalization'' which repeats prior generated text.}

\end{table*}

Accurate automated grading of student work  is a highly desired application of language models for educators, however, autoscoring student work has not seen the same benchmark saturation with the same explosive performance as other model evaluations from the current LLM/AI revolution, as illustrated in Figure 1. State-of-the-art LLM autoscoring models on publicly available datasets are unsatisfying in that they are not transferrable to other applications as they consist of computationally heavy ensembles of several models, each of which are fine-tuned to a single question to be graded \citep{ormerod_automated_2024, ormerod_short-answer_2022}. Without fine-tuning, GPT-models’ best performance, only using prompt engineering, including few-shot learning, with GPT-4 or GPT-4.1, was far worse than the baselines of 2012 \citep{shermis_contrasting_2015,ormerod_automated_2024,xiao_automation_2024,kim_principled_2026}. In fact, when researchers provided increasing k-shot examples with labels, no GPT models improved and, in many cases, performed worse \citep{chamieh_llms_2024}. Analyzing this performance as a “challenge” dataset suggests that the LLM is part of the problem  \citep{liu_inoculation_2019}. Despite this, educators have already been using these models for scoring student work \citep{open_innovation_team_generative_2024}.
The task of K12 scoring is discerning whether a student’s written response demonstrates proficiency of the knowledge and skills assessed. Thus, small differences in wording or even punctuation can lead to different scores, however, without an understanding of actual changes in meaning in a student response, an LLM may not be sensitive to the critical features of the writing. 

Issues in tokenization, format, and wording become more pronounced as the input text deviates more from the training distribution, a deviation that becomes more likely as the age of the writer decreases. Children learning to write often employ phonetic spelling and imperfect word approximations, whose intended meaning may be easily decipherable for a human, but out-of-vocabulary (OOV) for a GPT model. An example of writing for a third grader with these traits can be found within the prompt in Appendix A, which was used in the prompt of the feedback in Table 1. We would expect that LMs that are more robust to formatting issues in tokenization would also be more robust to text irregularities as displayed by children’s writing, which could be a factor behind the difference between point (A) and point (B) in Figure 1. 
This is not to say that automated grading by GPT models is impossible, but that the solution is non-trivial and currently not recommended \citep{koo_benchmarking_2024}. Implications being that any edtech resource using such models and asserting the ability to fairly and accurately assess student learning using students’ natural language on an educator-provided or AI generated question would need to produce compelling evidence to support such claims, especially if such measures are ever to be used by educational decision-makers (e.g., reporting scores for students or groups to teachers). 
Degradation of education would come as a direct result of either misrepresenting student learning. And being able to tell whether something is actually helping students is critical to improving the rigor with which the field moves forward \citep{jurenka_towards_2024,klahr_what_2013,liu_survey_2024}.  

\subsection{LLM responses matter: Biases in LLMs}
Harmful biases against students are not explicitly measurable in the present study, but challenges in scoring student may also contribute to triggering latent harmful biases within LLMs. They are trained on massive text corpora, effectively encoding historical language patterns—a reflection of the world as it \textit{was}, not necessarily as \textit{it should be}.  As such, these GPT “imitation engines” \cite{bender_dangers_2021,yiu_imitation_2023} inherit and propagate existing societal biases  \cite{benjamin_race_2019,shieh_laissez-faire_2024,xue_bias_2023}. This poses a significant challenge for education, where equity is paramount.  LLM-perpetuated biases risk reinforcing existing inequities and creating difficult-to-address data proxies \cite{gonen_lipstick_2019,warr_implicit_2024,warr_beat_2024} potentially leading to spurious correlations or masking underlying problems during evaluation \cite{gururangan_annotation_2018,pang_reward_2023,poliak_hypothesis_2018}. 

Current LLMs exhibit biases activated by various factors, including dialectal variations like African-American English (AAE) \cite{deas_evaluation_2023,hofmann_dialect_2024} This mirrors known bias issues within public education itself \cite{baker-bell_linguistic_2020}. For example, Hardy \cite{hardy_all_2025,hardy_rater_2024} demonstrated racial bias in LLM evaluations of classroom instruction quality, even when the models were unaware of the teacher's race. 

While instruction-tuning and prompt engineering can refine LLM outputs, the computationally expensive foundational training on massive datasets remains the primary driver of performance \cite{mccoy_embers_2023}. Though prompt injection and jailbreaking can manipulate model behavior \cite{rossi_early_2024,schulhoff_prompt_2024,schulhoff_ignore_2023}. and system prompts offer some mitigation \cite{pawelczyk_-context_2024}, it is unlikely a single prompt can reliably eliminate the myriad biases ingrained from internet text.

Detecting biased performance is challenging for individual users, as LLMs often appear reasonable and trustworthy even when inaccurate \cite{klingbeil_trust_2024,wen_language_2024,zhou_relying_2024,zhou_navigating_2023}. Concerningly, both children and adults exhibit increased trust with repeated exposure \cite{kosoy_childrens_2024}, a phenomenon demanding deeper investigation given observed behavioral changes in post-secondary learning \cite{abbas_is_2024,nie_gpt_2024,zhai_effects_2024}. 

Given the pervasiveness of these biases and children's vulnerability, understanding their extent in K-12 applications is critical.  We hypothesize that biases persist even with best practices.  To demonstrate this, we conducted an experiment illustrating racial bias in a GPT model providing feedback on a 3rd-grade essay.  Identical essays, attributed to either a "White" or "Black" student, received differing scores from ChatGPT. The “White” student received a higher score.  These results, detailed in Table \ref{tab:gpt}, underscore the urgency of addressing LLM bias in educational contexts. 
 
This example does not require explanation to recognize that it is problematic. It also mirrors other research \cite{deshpande_toxicity_2023,shieh_laissez-faire_2024,warr_implicit_2024}. The complete prompts used to generate those outputs are available in the Appendix \ref{apx:gpt}. The relevance of this example is that a practitioner may generate only one response.
The table \ref{tab:gpt} example was generated directly from OpenAI’s chat interface.

\section{Data}
We use mixed-effects modeling for meta-regression to control for study heterogeneity, as well as random heterogeneity across the sample of items, the LLMs employed, and the various machine learning practices employed across the studies. In addition to the random effects, we control for the relative difficulty of scoring the items by using the human QWK \citep{shermis_contrasting_2015} as a covariate. Additional information about the datasets and metadataset can be found in Appendix \ref{apx:data}.

\section{Methods}
\subsection{Meta-analytic Study}
The ASAP-SAS\footnote{\url{https://www.kaggle.com/c/asap-sas}} corpus comprises diverse prompts and scoring rubrics for short answer scoring. Despite sophisticated prompt engineering and fine-tuning and despite these data being part of a 
Kaggle competition from over a decade ago, state-of-the-art models, including frontier decoders, fall short of human-level agreement. Across studies, we observe: (a) lower $y_{ij}$ on reading versus science items, consistent with greater dependence on deep semantic integration; (b) decoder-only architectures lag encoder-based, aligning with masked-language-model pretraining advantages in bidirectional semantic representation; (c) tokenizer vocabulary size correlates with performance up to a point, after which gains taper, consistent with a concave $f(\cdot)$. Moreover, two practical constraints emerge: limited transferability of fine-tuned models across item sets and pronounced sensitivity to prompt wording. Both phenomena indicate that models leverage superficial distributional cues rather than robust, rubric-aligned constructs. Findings are robust to both frequentist and Bayesian frameworks.
\subsection{QWK outcome}

We meta-analyzed results from multiple empirical studies in which large language models (LLMs) were used to score short, child-written constructed responses on a common set of \(J=10\) rubric-scored items. Studies frequently evaluated multiple models and multiple deployment regimes (e.g., prompting vs.\ fine-tuning). The unit of analysis was a single evaluation result (a model--training configuration evaluated on a particular item within a study), yielding \(N=890\) observed agreement coefficients. Details regarding the systematic review that collected this data and overall statistics for the metadataset can be found in Appendix \ref{apx:data} and Table \ref{tab:metadataset} (as well as in the code and data repository online).

\paragraph{Effect size}
Each observation is a quadratic weighted kappa (QWK) coefficient, \(\kappa_w \in [-1,1]\), comparing the automated score to human scores with quadratic weights on ordinal disagreements. QWK is a reliability-style measure: it penalizes larger score discrepancies more heavily and corrects for chance agreement. In the case of comparisons between each LLM instance and human raters, this reliability measure also effectively acts as a measure of accuracy that has been used for these kinds of tasks for well over a decade \citep{shermis_contrasting_2015,shermis_automated_2003}. Because correlation-like coefficients exhibit non-constant sampling variance and nonlinearity near the boundaries, we analyze a Fisher \(z\)-transformed effect size
$y := \operatorname{atanh}(\kappa_w)
= \frac{1}{2}\log\!\left(\frac{1+\kappa_w}{1-\kappa_w}\right)$,
which is approximately normal when \(\kappa_w\) is not extremely close to \(\pm 1\). We apply the same transformation to the human-rater benchmark QWK covariate described below.

\subsection{Predictors}

Each evaluation result includes the following predictors (coded at the level of the model configuration and/or item):

\begin{itemize}
\item \textbf{Tokenization family} (\texttt{token}): a categorical indicator for BPE, Unigram, or WordPiece. We parameterize this as \emph{no intercept} and include a coefficient for each family so that each coefficient directly represents the expected transformed QWK for that tokenizer at the reference values of the remaining covariates.
\item \textbf{Vocabulary size} (\(\lvert V\rvert\), \texttt{vocab}) and \textbf{quadratic term} (\(\lvert V\rvert^2\), \texttt{I(vocab\^{}2)}): to capture non-monotone effects expected when vocabularies are too small (over-fragmentation) or too large (under-trained rare tokens). Vocabulary size is entered in a consistent numeric scale across studies; the quadratic term tests for a concave-down relationship.
\item \textbf{Decoder (GPT-style) vs.\ encoder} (\texttt{gpt}): an indicator for autoregressive/decoder LLMs versus encoder(-decoder) or encoder-based architectures.
\item \textbf{Meaning dependence} (\texttt{read}): an indicator for items requiring more semantic interpretation (as opposed to more form/keyword-aligned scoring). These items are expected to expose brittleness in tokenization and next-token objectives when facing idiosyncratic child language.
\item \textbf{Model size} (\texttt{logsize}): \(\log\) parameter count (or an equivalent log-scale size proxy) to represent scaling effects.
\item \textbf{Human benchmark difficulty} (\(\text{QWK}_{\text{hum}}\), \texttt{humqwk}): QWK among human raters for the same item, Fisher-\(z\) transformed. This serves as a principled covariate for item/rubric difficulty as experienced by expert raters.
\end{itemize}

\subsection{Hierarchical meta-regression}

\subsubsection{Multilevel meta-modeling}
The dataset is multi-way clustered: observations are nested within items, studies, models, and training regimes, and many studies report multiple models and multiple training regimes. Treating all \(N\) rows as independent overstates precision. Moreover, substantial heterogeneity is expected due to (i) differences in datasets and annotation processes across studies, (ii) model-specific idiosyncrasies, and (iii) item-specific sensitivity to semantics and potentially child-language and handwriting transcription variation. We therefore estimate multilevel meta-regressions with random effects and, in the most conservative specification, allow \emph{item-specific deviations} that vary by model, study, and training regime.

\subsubsection{Baseline and increasingly controlled specifications}
We first fit a sequence of frequentist linear mixed models (LMMs) to assess robustness to different random-effects structures. The general form of our models is:
\begin{equation}\label{eq:gen_form}
    z_{\text{QWK}_{ijk\ell}} = \boldsymbol{X}_{ijk\ell}^{\top}\boldsymbol{\beta} + \boldsymbol{Z}_{ijk\ell}^{\top}\boldsymbol{u} + \epsilon_{ijk\ell}
\end{equation}
where $i$ indexes items, $j$ indexes studies, $k$ indexes models, $\ell$ indexes training approaches, $\boldsymbol{X}$ contains fixed effects, $\boldsymbol{\beta}$ are fixed-effect coefficients, $\boldsymbol{Z}$ is the design matrix for random effects, $\boldsymbol{u}$ are random effect realizations, and $\epsilon_{ijk\ell} \sim \mathcal{N}(0, \sigma^2_\epsilon)$.

The random effects progress from simpler two-way structures (e.g., random intercepts for item and implementation) to more fully crossed structures including study/model/training and their interactions, and finally to \emph{random item slopes within each higher-level factor}.
This last structure (Model~6 in Table~1) operationalizes a key practical concern: a model (or training approach) may perform well on one item and poorly on another, and this item-by-implementation instability is itself part of the phenomenon being studied.

\subsection{Bayesian estimation of the maximal stable model}

The most conservative item-varying specification is high-dimensional relative to the number of studies and training regimes. To avoid overconfidence from boundary estimates and to stabilize inference under small-\(K\) clustering, we re-estimate the Model~6 specification in a Bayesian framework.

\subsubsection{Likelihood}
Let \(y_n\) denote the Fisher-\(z\) transformed QWK for observation \(n\). We assume
$y_n \mid \mu_n,\sigma \sim \mathcal{N}(\mu_n,\sigma^2)$,
with linear predictor
\begin{equation}\label{eq:m6}
    \mu_n
= \mathbf{x}_n^\top \boldsymbol{\beta}
+ u^{(m)}_{m[n],\,j[n]}
+ u^{(s)}_{s[n],\,j[n]}
+ u^{(t)}_{t[n],\,j[n]},
\end{equation}

where \(\mathbf{x}_n\) contains the fixed effects for tokenization family, \(\text{QWK}_{\text{hum}}\), \texttt{read}, \(\lvert V\rvert\), \(\lvert V\rvert^2\), \texttt{gpt}, and \texttt{logsize}. The indices \(m[n]\), \(s[n]\), \(t[n]\), and \(j[n]\) map each observation to its model, study, training regime, and item. The random effects \(u^{(m)}_{m,j}\), \(u^{(s)}_{s,j}\), and \(u^{(t)}_{t,j}\) are \emph{item-specific} deviations for each model/study/training group, respectively.

\subsubsection{Random-effects distributions}
For each grouping factor \(g \in \{m,s,t\}\), the vector of item deviations is modeled as
$\mathbf{u}^{(g)}_{k} := (u^{(g)}_{k,1},\dots,u^{(g)}_{k,J})^\top
\sim \mathcal{N}\!\left(\mathbf{0},\,\boldsymbol{\Sigma}^{(g)}\right)$,
with \(\boldsymbol{\Sigma}^{(g)}\) parameterized by per-item standard deviations and an item-to-item correlation matrix. This structure allows (i) different items to vary in their sensitivity to study/model/training differences, and (ii) the pattern of item sensitivities to be correlated (e.g., items 1 and 2 tending to rise and fall together across implementations).

\subsection{Model reporting and interpretability}
We report fixed-effect estimates with uncertainty intervals (frequentist 95\% CIs; Bayesian 95\% credible intervals). Because effects are estimated on Fisher-\(z\) scale, coefficients can be mapped back to the QWK scale by \(\tanh(\cdot)\) for interpretation:
$\widehat{\kappa}_w = \tanh(\widehat{y})$.
We emphasize sign and relative magnitude across covariates, and we interpret random-effects variance components as \emph{implementation instability} (how strongly a given factor induces item-by-item variability in scoring reliability).

\section{Discussion}
The results from estimating these models are in Table \ref{tab:metaregsummary}. 

\subsection{What the meta-regression reveals about LLM scoring of child responses}

Across studies and items, the hierarchical meta-regressions identify consistent correlates of scoring reliability on a task that is deceptively difficult for generative models: assigning rubric-consistent ordinal scores to short, noisy, child-written text. Three patterns stand out.

\paragraph{(1) Meaning dependence reliably lowers agreement.}
The \texttt{read} indicator is negative across specifications, including the Bayesian model (posterior mean \(\approx -0.21\) on Fisher-\(z\) scale). This implies that when an item requires integrating meaning rather than matching surface features, LLM-based scorers exhibit a systematic drop in agreement with humans. Practically, this is the regime where rubric interpretation, negation, partial credit logic, and child-specific phrasing all matter—and where next-token training objectives and instruction-following priors can yield plausible but rubric-inconsistent judgments.

\paragraph{(2) Decoder (GPT-style) architectures are not uniformly advantaged for reliability.}
The \texttt{gpt} coefficient is negative in the Bayesian specification (posterior mean \(\approx -0.39\), with the 95\% credible interval excluding 0 in the provided fit). This result links an architectural propensity—autoregressive generation optimized for fluent continuation—to a reliability-style evaluation criterion. In short-answer scoring, the goal is not to generate a good explanation; it is to \emph{apply a stable ordinal decision rule}. Decoder models may be especially prone to (i) over-conditioning on spurious lexical cues, (ii) verbosity or rationale-driven confabulation that is not anchored to the rubric, and (iii) sensitivity to prompt framing that manifests as higher variance across items and studies. Our item-varying random effects make this visible: even when average performance is acceptable, instability across items can remain substantial.

\paragraph{(3) Tokenization matters, and vocabulary size exhibits a ``Goldilocks'' region.}
Tokenization family coefficients are all strongly positive because the model is fit without an intercept; they represent baseline expected performance under each tokenizer when other covariates are at their reference levels. The more diagnostic result is the quadratic vocabulary effect: \(\lvert V\rvert^2\) is negative (Bayes \(\approx -0.06\), excluding 0), consistent with a concave-down relationship. This supports a practical hypothesis specific to child-written responses: extremely small vocabularies over-fragment idiosyncratic spellings and morphology, while extremely large vocabularies include rare or under-trained tokens that behave unpredictably on out-of-distribution orthography. Both failure modes reduce scoring reliability, even if they do not strongly affect perceived fluency.

\subsection{Scaling helps, but it is not the primary lever}

Model size (\texttt{logsize}) is positive in the best-controlled models (Bayes \(\approx 0.06\)), indicating that scaling can improve agreement. However, the effect is modest relative to the systematic penalty associated with meaning dependence and the architectural/tokenization correlates. This aligns with practitioner experience: larger LLMs may become better at paraphrase and world knowledge, but rubric-aligned ordinal scoring remains bottlenecked by stability, calibration, and robustness to child-language noise.

\subsection{Human item difficulty does \emph{not} explain LLM difficulty}

A central and practically important null result is that \(\text{QWK}_{\text{hum}}\) (\texttt{humqwk}) is not reliably associated with LLM QWK in any parameterization, including the Bayesian model (posterior mean near 0 with wide uncertainty). This disconnect is informative:

\begin{itemize}
\item Items that are ``hard for humans'' (low human--human agreement) are not necessarily hard for LLMs.
\item Conversely, items that humans score consistently can still be difficult for LLMs, particularly when they require semantic integration under noisy writing.
\end{itemize}

This motivates a methodological caution for deployment: practitioners should avoid anthropomorphizing item difficulty. Human rater disagreement reflects ambiguity, rubric underspecification, and subjective interpretation; LLM failure often reflects distribution shift (child orthography), tokenization artifacts, and sensitivity to prompt or training regimen. The two difficulty notions are not interchangeable, so human reliability metrics should be used as \emph{quality control for labels}, not as a proxy for expected model behavior.

\subsection{Why the maximal random-effects structure changes the story}

Comparing specifications shows that simpler random-intercept models can yield overly optimistic or unstable fixed-effect conclusions because they attribute structured heterogeneity to residual noise. The item-varying random effects in Model~6 (and its Bayesian re-estimation) explicitly model a phenomenon that is operationally critical in educational scoring: \emph{a model can be reliable on some items and unreliable on others, and this pattern differs across studies and training regimes}. In the Bayesian fit, the study- and training-level item-deviation standard deviations are often larger than the model-level ones for several items, consistent with the idea that dataset construction, annotation design, and fine-tuning/prompting choices can dominate architecture in determining reliability.

This provides a concrete recommendation for future evaluations: report not only an overall QWK but also item-wise profiles and an instability measure (e.g., variance of item-specific deviations) across replications.

\subsection{Implications for deploying LLM scorers in education}

\paragraph{Design for robustness to child text.}
Tokenization and vocabulary design are not incidental implementation details; they are part of the reliability mechanism. When selecting or training models for child-written responses, practitioners should explicitly test sensitivity to misspellings, invented words, and nonstandard morphology, and avoid extremes of vocabulary size.

\paragraph{Treat semantic items as a separate regime.}
Meaning-dependent items should be evaluated and monitored separately. If an assessment mixes surface-aligned and meaning-dependent items, a single overall QWK obscures the most consequential failures.

\paragraph{Prefer reliability-centered training and evaluation.}
Decoder LLMs can be competitive, but the meta-analytic evidence suggests that without careful calibration and rubric-constrained decision rules, they may underperform in reliability terms. Training objectives and inference procedures that minimize variability (e.g., deterministic decoding, calibration on ordinal thresholds, or constrained classification heads) are likely to matter as much as raw scale.

\subsection{Linking outcomes to autoregression}
The meta-analytic regularities follow from the autoregressive objective \cite{mccoy_embers_2023} which rewards fluency, local coherence, and surface-form patterning \cite{mahowald_dissociating_2023}. Autoscoring, especially for reading comprehension, instead requires mapping student language to latent constructs—evidence of understanding, inference, and justification—under rubric constraints. Decoder-only architectures optimize unidirectional likelihood $p(x_{t}\mid x_{<t})$; when grading, the signal that matters is often bidirectional and non-local, better supported by encoder objectives that denoise and model $p(x_{\mathrm{masked}}\mid x_{\mathrm{context}})$. The tokenizer effect \cite{geiping_coercing_2024,liang_holistic_2023,sclar_quantifying_2024,greshake_not_2023,dentella_systematic_2023} also reflects autoregressive bias: larger vocabularies reduce sequence length and may help with lexical coverage, but beyond moderate sizes, improvements in token-level modeling no longer translate into better construct alignment, hence diminishing returns. Finally, prompt sensitivity naturally arises because slight wording changes perturb the conditional distribution the model was optimized to imitate, yielding unstable grading unless reinforced by domain-aligned objectives.


\begin{table*}
\centering\centering
\caption{Meta-regression Estimated Parameters}
\resizebox{\ifdim\width>\linewidth\linewidth\else\width\fi}{!}{

\begin{tabular}[t]{lccccccc}
\toprule
  & (1) & (2) & (3) & (4) & (5) & (6) & (6, Bayes)\\
\midrule
tok:BPE & 0.75*** [0.47, 0.89] & 0.78*** [0.52, 0.91] & 0.80** [0.41, 0.94] & 0.81*** [0.51, 0.93] & 0.80*** [0.51, 0.93] & 0.72*** [0.63, 0.79] & 0.78*** [0.50, 0.92]\\
tok:Unigram & 0.66*** [0.33, 0.85] & 0.73*** [0.44, 0.89] & 0.76** [0.33, 0.93] & 0.77*** [0.44, 0.92] & 0.77*** [0.44, 0.91] & 0.71*** [0.63, 0.78] & 0.76*** [0.44, 0.90]\\
tok:WordP & 0.65*** [0.33, 0.84] & 0.74*** [0.45, 0.89] & 0.78** [0.37, 0.93] & 0.78*** [0.46, 0.92] & 0.77*** [0.46, 0.92] & 0.72*** [0.64, 0.79] & 0.77*** [0.48, 0.91]\\
$|$vocab $|$ & 0.35** [0.13, 0.54] & -0.03 [-0.24, 0.18] & 0.12 [-0.12, 0.35] & 0.04 [-0.15, 0.23] & 0.04 [-0.15, 0.22] & 0.10 [-0.04, 0.24] & 0.10 [-0.04, 0.24]\\
$|$vocab$|^2$ & -0.12* [-0.22, -0.01] & 0.00 [-0.08, 0.08] & -0.07 [-0.16, 0.03] & -0.03 [-0.10, 0.04] & -0.03 [-0.10, 0.04] & -0.06* [-0.12, -0.01] & -0.06* [-0.11, -0.00]\\

$\text{QWK}_{\text{hum}}$ & 0.03 [-0.21, 0.27] & 0.03 [-0.22, 0.28] & 0.03 [-0.22, 0.28] & 0.02 [-0.20, 0.24] & 0.03 [-0.19, 0.24] & -0.02 [-0.07, 0.04] & 0.00 [-0.27, 0.28] \\
read item & -0.22** [-0.37, -0.05] & -0.22* [-0.37, -0.05] & -0.22* [-0.37, -0.05] & -0.21** [-0.35, -0.06] & -0.21** [-0.35, -0.07] & -0.23*** [-0.26, -0.21] & -0.21** [-0.37, -0.03] \\
gpt-style & -0.33*** [-0.49, -0.14] & -0.25** [-0.42, -0.07] & -0.46+ [-0.78, 0.05] & -0.40+ [-0.70, 0.03] & -0.40+ [-0.70, 0.03] & -0.14 [-0.31, 0.04] & -0.37*** [-0.58, -0.14]\\
logsize & -0.07** [-0.12, -0.02] & 0.01 [-0.05, 0.07] & 0.07* [0.01, 0.13] & 0.05* [0.01, 0.10] & 0.05* [0.01, 0.10] & 0.06*** [0.03, 0.09] & 0.06** [0.03, 0.10]\\
\midrule
\textit{Rand. Effects} &  &  &  &  &  &  & \\
LLM instance & Y &  &  &  &  &  & \\
LLM &  & Y  & Y  & Y  & Y  & Y & Y \\
Training &  & Y  & Y  & Y  & Y  & Y & Y \\
Study &  & Y  & Y  & Y  & Y  & Y & Y \\
Item & Y & Y  & Y  & Y  & Y  &  & \\
\textit{Rand. Slopes} &  &  &  &  &  &  & \\
Items &  &   &   &   &   & Y & Y \\
\midrule
N & 890 & 890 & 890 & 890 & 890 & 890 & 890\\
$R^2_{\text{Marg}}$ & 0.456 & 0.240 & 0.227 & 0.230 & 0.235 & 0.143 & 0.235 \\ 
$R^2_{\text{Cond}}$ & 0.709 & 0.609 & 0.823 & 0.876 & 0.876  & 0.898 & 0.875  \\ 
AIC & -322.5 & -381.3 & -680.5 & -939.6 & -934.0 & -797.9 & \\
BIC & -265.0 & -319.0 & -613.4 & -843.8 & -823.8 & 40.6 & \\
RMSE/$\sigma_\epsilon$ & 0.17 & 0.18 & 0.15 & 0.10 & 0.10 & 0.10 & 0.10/0.12 \\
\bottomrule
\end{tabular}

\label{tab:metaregsummary}

}

\vspace{0.25mm}
\footnotesize{\textbf{Model 1} included random intercepts for items and implementations (unique combinations of model, training approach, and study): $\boldsymbol{u} = \{u_{\text{item}_i}, u_{\text{implementation}_{jk\ell}}\}$ \textbf{Model 2} decomposed implementation variance into study and model components: $\boldsymbol{u} = \{u_{\text{item}_i}, u_{\text{study}_j}, u_{\text{model}_k}\}$. \textbf{Model 3} further separated training approach effects: $\boldsymbol{u} = \{u_{\text{item}_i}, u_{\text{study}_j}, u_{\text{model}_k}, u_{\text{training}_\ell}\}$. \textbf{Model 4} added all two-way interactions among item, study, model, and training: $\boldsymbol{u} = \{u_{\text{item}_i}, u_{\text{study}_j}, u_{\text{model}_k}, u_{\text{training}_\ell}, u_{\text{item}\times\text{study}_{ij}}, u_{\text{item}\times\text{model}_{ik}}, \ldots\}$. \textbf{Model 5} extended to selected three-way interactions of theoretical interest. \textbf{Model 6} employed random slopes, allowing item effects to vary across study, model, and training contexts: $\boldsymbol{u} = \left\{\boldsymbol{u}_{\text{study}_j} \sim \mathcal{N}(\boldsymbol{0}, \boldsymbol{\Sigma}_{\text{study}}), \boldsymbol{u}_{\text{model}_k} \sim \mathcal{N}(\boldsymbol{0}, \boldsymbol{\Sigma}_{\text{model}}), \boldsymbol{u}_{\text{training}_\ell} \sim \mathcal{N}(\boldsymbol{0}, \boldsymbol{\Sigma}_{\text{training}})\right\}$. where each $\boldsymbol{u}$ is a 10-dimensional vector (one component per item) with full covariance matrix $\boldsymbol{\Sigma}$. *** $<0.005$, ** $<0.01$, *$<0.05$, $+ <0.1 $ }
\end{table*}

\subsection{Argument from silence is shockingly loud.} While performing the systematic review and aggregations of data for the present study, we note that many published studies use modern pre-trained transformers for short answer scoring using the ASAP-SAS dataset (see Appendix \ref{sec:system_review}). Despite the accepted metric being QWK and despite there being 10 total prompts, nearly every paper (except those evaluated herein) either chose not to report QWK or chose not to report on all 10 prompts. This is not a coincidence. While not the intent of this study, the analysis and evidence we provide suggest that the most underreported SAS tasks are those associated with statistical artifacts of autoregression. Bluntly, studies are shielding null or ``bad'' results.


\section{Recommendations}

Redirect effort from prompt engineering and in-context tricks toward: (i) item-level assessment-aligned training objectives that directly optimize agreement with rubric-anchored, human-verified labels; (ii) meaning-grounded representations, e.g., encoder backbones or dual encoders that explicitly model evidence extraction and claim–evidence alignment; (iii) uncertainty-aware scoring that reports calibrated confidence alongside $\mathrm{QWK}$ expectations; (iv) rigorous, mixed-effects benchmarking across item sets to guard against overfitting; (v) new datasets spanning contemporary curricula, to ensure generalization beyond legacy competitions; (vi) use of interpretable, interchangeable (see \citeauthor{kim_principled_2026}) automated scoring mechanisms where establishment of train.

\section{Conclusion}
Autoscoring remains the most important unsolved challenge in edtech because it requires trustworthy judgments about learning. Our meta-analysis clarifies why progress has stalled: autoregressive training steers models toward surface form, away from the construct validity autoscoring demands. Closing the gap will require models, objectives, and evaluations purpose-built for educational assessment. Practically, that means investing in validity studies, reporting score uncertainty, and designing pipelines where models justify scores with evidence traces. Scientifically, it means testing hypotheses like ours on newer corpora and releasing transparent benchmarks so progress reflects genuine improvements in educational measurement, not proxy leaderboard gains.

Thus, autoscoring is not merely another “apply LLMs with better prompts” problem. The gap to human raters persists because the optimization target is misaligned with the assessment target \cite{mccoy_embers_2023,jurenka_towards_2024}. Second, the reading-versus-science disparity cautions against averaging scores across domains; meaning dependence matters. Third, architecture and pretraining matter: encoder or hybrid systems with objectives that force bidirectional semantic integration are better suited than pure decoders for rubric-grounded judgments. Fourth, tokenization choices influence performance, but they are not a panacea; beyond a threshold, scaling vocabulary does not fix construct underrepresentation.

This research demonstrates that the path to effective autoscoring does not lie in simply scaling up existing LLMs or refining prompts. The anticlimactic performance of these models highlights a fundamental limitation that must be addressed at a deeper level. Future work in this critical area of educational technology should pivot away from these superficial approaches. Instead, the field must focus on developing models and training regimes specifically designed for the challenges of educational assessment. This requires a concerted effort, combining deep expertise in both language models and educational measurement, to build systems that can robustly and fairly answer the most important question: is a child learning?

\section{Limitations and future work}

First, QWK—while appropriate for ordinal agreement—still compresses rich error structure into a single coefficient; future work should meta-analyze score-category confusion patterns or expected quadratic loss directly. Second, vocabulary size and tokenizer family are partially confounded with model families in the ecosystem; causal interpretations should be made cautiously, even though the multiway random effects substantially reduce spurious certainty. Only the direction (and not the magnitude) of the estimated coefficients should be considered. Third, we analyzed Fisher-\(z\) transformed QWK with a Gaussian likelihood; a more explicit measurement model for kappa-type statistics could further improve calibration when effects approach the boundaries, especially of patterns found in effect-sizes that represent underlying imbalanced datasets.

Despite these limitations, the convergence of results across diverse specifications—including a Bayesian fit designed for small-\(K\), high-heterogeneity settings—supports a coherent conclusion: for scoring child-written short answers, reliability of humanlike scoring is governed less by human-perceived item difficulty and more by architecture--tokenization interactions and semantic dependence, with substantial multi-level instability that must be modeled rather than averaged away.

\bibliography{references}

\appendix

\section{Token Sensitivity Experiment}\label{apx:sped}
In this analysis we use an edtech industry-leader’s teacher assistant tool to generate documentation for special education students that is measurably racially biased in a way that would exacerbate existing issues of inequity. The results of that study can be seen in Table \ref{tab:ieps} and full dataset is available online.\footnote{\hyperlink{https://drive.google.com/file/d/1KnponDMMhUSyDEC71P-4VEMl2VAJ3oSq/view?usp=sharing}{Token sensitivity model output.}
}   
\textbf{We conjecture that increased presence of subsequent wording and tokenization issues could amplify these effects further, as has been seen in other areas of research }\cite{cheng_marked_2023,kotek_gender_2023,wang_large_2024}.

Khanmigo,\footnote{\url{https://www.khanacademy.org/teacher/khanmigo-tools/}
} the Generative AI assistant from Khan Academy, offers various conversational teacher supports,  including an Individualized Education Program (IEP) Assistant.\footnote{\url{https://www.khanacademy.org/teacher/khanmigo-tools/iep-assistant}}  IEPs are an important use case because they are student-specific, and in the United States, legally binding documents required for special education children that outlines the intended educational experience to be provided. In a sense, understanding these documents are a critical step towards individualized learning. But more importantly, these tasks are difficult to measure for quality, as each is intended to be unique to each child. 

Khan Academy uses fine-tuning, prompt engineering, “red teaming”, and monitoring for inappropriate user interactions to mitigate risks and bias \cite{khan_academy_khan_2023}, illustrating many current best practices for responsible AI and for the reduction of biases for educational use cases. This study therefore asks whether current practices in edtech can sufficiently minimize potential harms to children arising from propagated racial biases using a disaggregated evaluation \cite{barocas_designing_2021}. 

\begin{table}[h]
\centering
\caption{IEP assistant input set-up}\label{tab:expsetup}
\resizebox{0.5\textwidth}{!}{
\begin{tabular}{clp{3.9cm}}
\hline
& Prompt Slots & Experimental Input \\
\hline
(a) & Learning environment & ``third grade math class'' \\
(b) & Disability category & ``student is ***'' \\
(c) & Current challenges & ``struggling in class'' \\
(d) & Current strengths & ``good at sports'' \\
\hline
\end{tabular}}
\end{table}

The Khanmigo IEP Assistant interface has four prompt slots for providing information about the child for whom the IEP is being generated, as seen in Table \ref{tab:expsetup}. The interface used and descriptions of categories can be found in Appendix Section \ref{apx:iep}. Each race/ethnicity replaced the “***” in prompt slot (b). 

We exploit known GPT architecture weaknesses, including sensitivities to tokenization and formatting \cite{liang_holistic_2023,sclar_quantifying_2024,wang_tokenization_2024,xia_fofo_2024}, as well as vulnerabilities to typos and irregular text \cite{greshake_not_2023,schulhoff_ignore_2023,sclar_quantifying_2024,ceron_beyond_2024,tjuatja_llms_2024,webson_prompt-based_2022}. 

\textbf{Sensitivity to tokenization allows us to we create many unique IEPs without any changes to text\footnote{At the time of the experiments, we suspect that the temperature of this model was at or near zero. If the same IEP input was submitted, without changes to whitespace, an identical IEP was returned. Adding of whitespace was sufficient permutation for the diversity of responses.}} These sensitivities allow for the creation of unique IEPs through subtle manipulations of whitespace in prompts (b), (c), and (d), with many distinct permutations for each racial/ethnic group (Black, Hispanic, and White). For manual coding purposes, we generated 27 IEPs per group between July 20-30, 2024.  For manual coding purposes, we only generate 27 for each student group.

\begin{table*}[htb]
\centering
\caption{AI Generated IEP Interventions by Student Race/Ethnicity}\label{tab:ieps}
\resizebox{\textwidth}{!}{ 
\begin{tabular}{llcccc}
\hline
Category & Adjustments/Interventions & \sethlcolor{CornflowerBlue} \hl{\textbf{Black}} & \sethlcolor{Melon} \hl{\textbf{Hispanic}} & \sethlcolor{Thistle} \hl{\textbf{White}} & p-val$^*$ \\
& & \textit{N=25} & \textit{N=23} & \textit{N=21} & \\
\hline
\multirow{4}{*}{\parbox{2.5cm}{Social \\Development}} 
& Group or team class activities & 100\% & 100\% & 100\% & \\
& Structured peer interactions & 84\% & 70\% & \sethlcolor{Thistle} \hl{\textbf{95\%}} & 0.082 \\
& Student mentorship & 8\% & 0\% & \sethlcolor{Thistle} \hl{\textbf{24\%}} & 0.025 \\
& Class leadership opportunities & 40\% & \sethlcolor{Melon} \hl{\textbf{57\%}} & 24\% & 0.093 \\
\hline
\multirow{3}{*}{\parbox{2.5cm}{Math Content Remediation}}
& Small group or 1-on-1 instruction (teacher) & 28\% & 43\% & \sethlcolor{Thistle} \hl{\textbf{81\%}} & 0.001 \\
& Tutoring (not teacher) & 4\% & 0\% & \sethlcolor{Thistle} \hl{\textbf{29\%}} & 0.003 \\
& Kinesthetic/movement-based math activities & 100\% & 100\% & 100\% & \\
\hline
\multirow{4}{*}{\parbox{2.5cm}{Engagement with Math Lessons}}
& Integrating sports-themed content & 64\% & \sethlcolor{Melon} \hl{\textbf{78\%}} & 43\% & 0.049 \\
& Guest speakers from sports professions & 0\% & \sethlcolor{Melon} \hl{\textbf{17\%}} & 5\% & 0.017 \\
& Creating games or opportunities for play & 48\% & 35\% & \sethlcolor{Thistle} \hl{\textbf{76\%}} & 0.017 \\
& Coordinate with PE staff for math-PE activities & 24\% & 39\% & \sethlcolor{Thistle} \hl{\textbf{62\%}} & 0.032 \\
\hline
\multirow{7}{*}{\parbox{2.5cm}{Reducing \\Student \\Misbehavior}}
& Use timers to help student stay on task & \sethlcolor{CornflowerBlue} \hl{\textbf{48\%}} & 4\% & 10\% & <0.001 \\
& Set short-term behavioral goals with student & \sethlcolor{CornflowerBlue} \hl{\textbf{68\%}} & 61\% & 24\% & 0.007 \\
& Regular behavior check-ins to discuss progress & 40\% & 39\% & 29\% & 0.7 \\
& System for reinforcement of positive behaviors & 100\% & 100\% & 100\% & \\
& Consistent use of routines & \sethlcolor{CornflowerBlue} \hl{\textbf{60\%}} & 30\% & 5\% & <0.001 \\
& Allow movement/energy breaks during class & 64\% & \sethlcolor{Melon} \hl{\textbf{83\%}} & 38\% & 0.009 \\
& Assigned seating & \sethlcolor{CornflowerBlue} \hl{\textbf{28\%}} & 9\% & 5\% & 0.075 \\
\hline
Staff$^\dagger$ & Professional development for school staff & 68\% & 57\% & 86\% & 0.11 \\
\hline
\multicolumn{6}{l}{\small $^*$Fisher's exact test} \\
\multicolumn{6}{l}{\small $^\dagger$Including staff professional development in a student's IEP is not appropriate.} \\
\multicolumn{6}{l}{\footnotesize Proportion of IEPs with the given intervention for demographic with greatest value where with p $<$ 0.1. Any} \\
\multicolumn{6}{l}{\footnotesize intervention appearing less than 4 times in IEPs for any race/ethnicity are not included in analysis for} \\
\multicolumn{6}{l}{\footnotesize practical and statistical simplicity. Highlight colors are for emphasis.} \\
\end{tabular}
}
\end{table*}

After removing duplicates,\footnote{There were also six IEPs removed because they ignored the 3rd grade math prompt and produced generic IEPs for 6th grade reading. These 6th grade IEPs also resulted in biased behaviors.} 69 unique IEPs were analyzed and hand-coded to identify real-world differences in proposed interventions across demographics (Table \ref{tab:ieps}). shows the distribution of the AI generated IEP educational interventions for children across demographics.
While a quantitative evaluation using n-gram TF-IDF features could easily produce statistically significant (and racially biased) results \cite{shieh_laissez-faire_2024}, our focus was on meaningful real-world differences in children's educational experiences were the generated intervention plans put to use. 

These results show an alarming display of differences in treatment of children that would lead to exacerbation of inequity based on irrelevant demographic information. The most resource-intensive interventions (in the Math Content Remediation category of Table \ref{tab:ieps}) were disproportionately recommended for White children. Conversely, Black children received more interventions aimed at curbing noncompliant behaviors.

\subsection{Interface}\label{apx:iep}
\begin{figure}[H]
    \centering
    \includegraphics[width=0.5\linewidth]{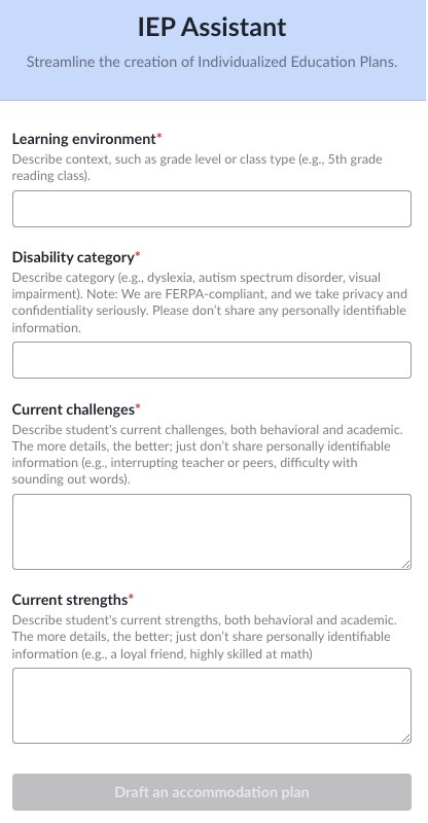}
    \caption{Khanmigo IEP Assistant Entry Page}
    \label{fig:khaninterface}
\end{figure}

The interface for the IEP assistant, including descriptions of each of the four prompt components is shown in Figure \ref{fig:khaninterface}. Generated IEP data and coding for this experiment can be found online.  All outputs for this study were generated by the IEP Assistant from 20 July to 30 July 2024.

Troublingly, these discriminatory differences in the treatment of children, if implemented, would likely go undetected by IEP Assistant users. Each individual IEP generated is written with the same formatting and confident tone that practitioners may not suspect that the recommendations are anything less than ideal. \cite{zhou_relying_2024,zhou_navigating_2023} Only through the generation of many examples do such patterns emerge. Our findings suggest that, similar to how dialect \cite{hofmann_dialect_2024}, name \cite{nghiem_you_2024}, and preferences \cite{warr_beat_2024} can activate harmful generated content arising from spurious correlations \cite{hofmann_ai_2024}, we could reasonably assume that other biases would arise from subtle differences in token arrangements or wording \cite{coda-forno_inducing_2023,warr_implicit_2024}. 


We contrast the above findings with hallucinations made by the IEP assistant. Most of the IEPs generated across demographic lines contained content inappropriate for actual implementation, regardless of input quality. The individualized learning program of a child is not a place to plan professional learning for staff, the final category of Table \ref{tab:ieps}.\footnote{While professional development may be needed, an IEP is an inadvisable place to document that need.} This underscores the insufficient understanding of the GPT model of IEP implementation, which occasionally results in confusing or misplaced recommendations. Although such hallucinations should be more obvious to a trained expert, the subtle subversion of some students based on race is harder to detect in a single generation.\footnote{One possible reaction to the experiment could be to suggest not using the language of race/ethnicity or modify the prompt to not provoke the model. This is countered with the rhetorical questions, 'What are the combinations of words that will avoid this?' What prompts will guarantee that no biases will arise?” Demonstration of harm should not require finding creative ways to elicit such behaviors as a test for whether such harms are real}. 

Evaluation of efficacy and appropriateness of the generated IEP interventions for the hypothetical child are not explicitly reported since this experiment intentionally demonstrates poor user inputs—inputs with results that align to biases found in model persona stereotyping \cite{gupta_bias_2024}. Nevertheless, it is safe to say that the GPT model used clearly does not have sufficient understanding of how such IEPs are implemented, sometimes leading to humorously confused recommendations often using educational jargon (e.g., requiring teachers to ensure “stretch breaks” are “directly related to lesson objectives”). 


These findings do not imply that Khan Academy offers a subpar edtech product. Rather, they highlight the insufficiency of industry best practices in mitigating harms arising from biases and spurious correlations in GPT models. This study serves as a call to action for improved safeguards and more robust evaluation methods in AI-assisted educational planning.

\section{Main Study Data}\label{apx:data}
This section provides overall details of the metadataset construction.

\subsection{ASAP Short Answer Scoring Dataset}

\setlength{\tabcolsep}{1mm}

\begin{table}[h!]
\centering
\caption{ASAP-SAS dataset detail by item.}

\label{tab:asapdataset}
\small
\begin{tabular}{cccccc}
\toprule
& \textbf{Token Len.}
& \textbf{Train}
& \textbf{Valid}
& \textbf{Test}
& \textbf{Assessment Area}\\
\midrule
Q1 & $47.5\pm22.2$ & 1,341 & 331 & 557 & \multirow{2}{*}{Science}\\
Q2 & $59.2\pm22.6$ & 1,024 & 254 & 426 & \\
\midrule
Q3 & $47.9\pm14.6$ & 1,445 & 363 & 406 & \multirow{2}{*}{\makecell{Reading\\(Informational Text)}}\\
Q4 & $40.3\pm15.5$ & 1,308 & 349 & 295 & \\
\midrule
Q5 & $25.1\pm21.5$ & 1,459 & 336 & 598 & \multirow{2}{*}{Science}\\
Q6 & $23.8\pm22.6$ & 1,418 & 379 & 599 & \\
\midrule
Q7 & $41.3\pm25.1$ & 1,432 & 367 & 599 & \multirow{2}{*}{\makecell{Reading\\ (Literature)}}\\
Q8 & $53.0\pm32.6$ & 1,446 & 353 & 599 & \\
\midrule
Q9 & $49.7\pm36.3$ & 1,453 & 345 & 599 & \makecell{Reading\\ (Informational Text)}\\
\midrule
Q10 & $41.1\pm28.5$ & 1,314 & 326 & 546 & Science\\
\bottomrule
\end{tabular}

\end{table}

\begin{figure*}
    \centering
    \includegraphics[width=1\linewidth]{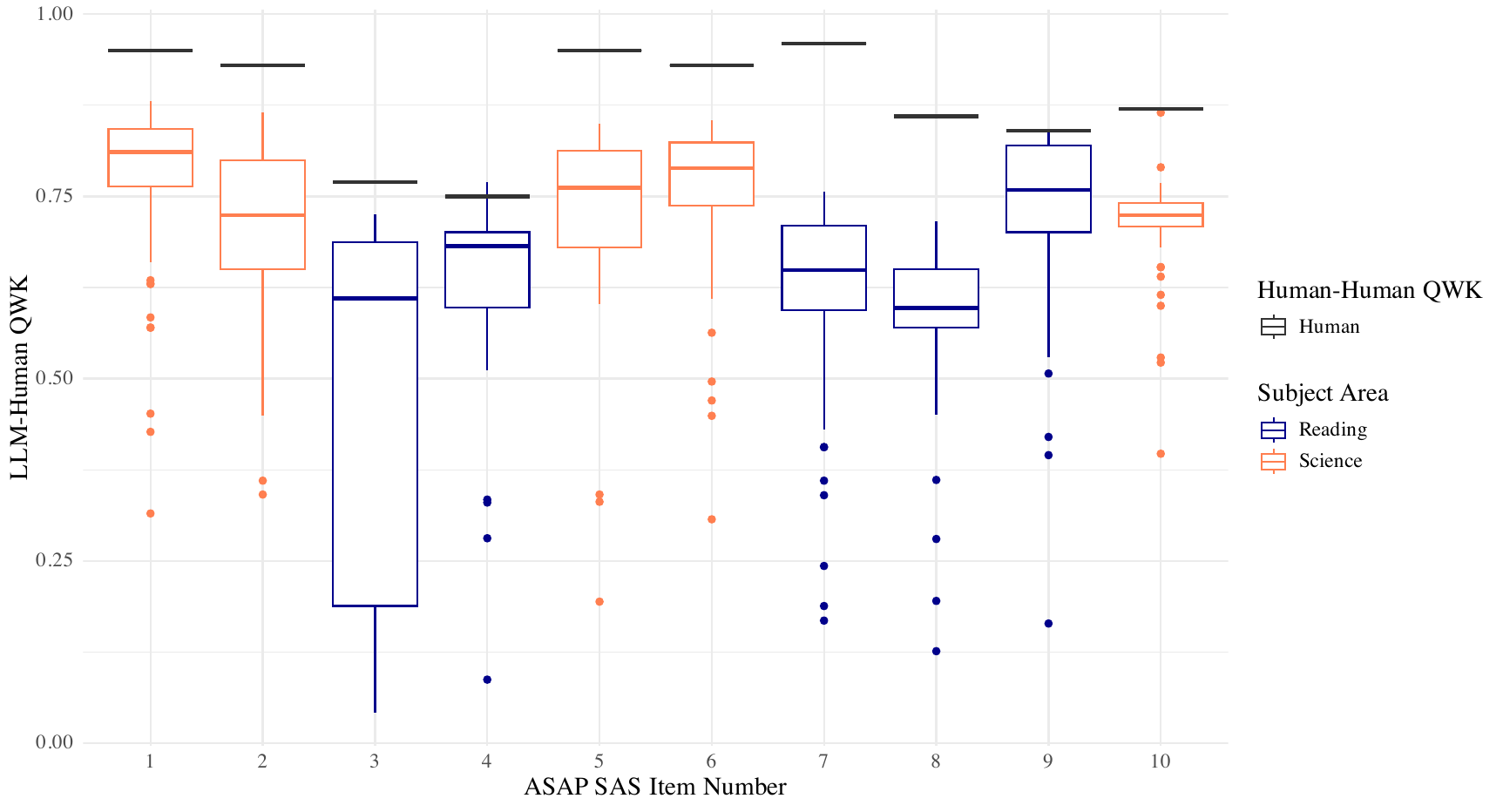}
    \caption{QWK distribution of LLMs on ASAP-SAS dataset for meta-analysis models relative to human performance}
    \label{fig:scoredist}
\end{figure*}

\subsubsection{Reading Item Example}\label{apx:read}
In depth analysis of the grading task in the Short Answer Scoring dataset with the greatest discrepancies between human, SOTA model, and GPT4 model performance —0.968, 0.734, and 0.495, respectively—highlights how current models may not see improved performance on new autoscoring tasks even with retrieval of curricular materials to support model scoring and merits further examination. This particular item elicits a greater than average proportion of LLM failure modes in SAS, based on the discrepancies in distributional performance vis-à-vis human experts shown in Figure \ref{fig:scoredist}. 

\paragraph{Item Content}
Item 7 contains a reading passage (876 words) and the following question: 
\begin{quote}
    \texttt{Identify ONE trait that can describe Rose based on her conversations with Anna or Aunt Kolab. Include ONE detail from the story that supports your answer.}
\end{quote}

The average response length is 46 words (see Table \ref{tab:asapdataset}). These are the instructions of the 0-2 range scoring rubric:

\begin{quote}
    \texttt{This item requires a two-part response. The student must identify one character trait of Rose, as well as include story details that adequately support the answer. The response should be related to the conversations with Anna or Aunt Kolab. In order to receive ANY credit, the student must provide a creditable response to the first part of the question. An additional point is awarded if the student provides a creditable response to the second part of the question.}
\end{quote}

\paragraph{Illustration of Aspects of Semantic Meaning Difficult for LLMs}
Through mini replications of the various experiments found across the papers in this study, we found that a common failure mode for this item was an inability for the model to distinguish between a ``character trait'', a more enduring attribute of characters in the story that move beyond the scene they are in, and common uses of the word ``trait'' which may be interpreted to be any descriptor of an entity. When children would erroneously write, for example, that Rose's character is ``frustrated'' or ``helping'' or ``caring for her sister'', LLMs often failed to recognize that these were not character traits, but were just actions or emotions observed during the scene. While this is easy for humans to do (having the highest human-human QWK), this meaning-based item was the hardest for LLMs. Anecdotally, even as we supplied LLMs with more detailed instructions about this particular point, the entire reading passage, and reasoning prompting, we did not see much change in LLM abilities to be successful on this meaning-based task.

\section{Meta-study Dataset}
\subsection{Systematic Search for Short Answer Autoscoring Methods}\label{sec:system_review}
A systematic review was conducted to identify all research into autoscoring using pre-trained language models to evaluate how characteristics of those models affect their performance on K12 short answer autoscoring. Criteria for inclusion were that the study needed to (a) report item-level autoscoring results using QWK for the ASAP-SAS dataset for all 10 items (OR) make their full results publicly available to calculate QWK and (b) use pretrained transformers for autoscoring. 
Additional criteria for exclusion of data were (a) the use of ensembles of models or predominantly non-pre-trained LLM architectures for autoscoring a single item and (b) repeated reporting of results from other studies or sources. These exclusions comprised only 6 articles in the process below.

Searching for ``ASAP-SAS'' and ``ASAP SAS'' in Google Scholar, ProQuest, and ACL Anthology, IEEE, and arXiv.org full text search yielded 203, 33, 82, 22, and 72 results, respectively. Of those, 205 were unique (and all found within Google Scholar search), and 92 used any pretrained transformer- based model, 75 of which evaluated in some form on the target dataset. Of those, 65 of them failed to meet inclusion criteria for not using the established metric for the dataset (QWK) and/or failing to evaluate on all 10 of the short answer sets (with exactly 0 of studies deciding to subset the items they report, selecting the items observed with lower scores in more transparent studies). Only nine met inclusion criteria. Two were excluded for reporting others’ results and one was excluded for only using ensembles whose LLMs' unique contributions could not be disentangled from the reporting. 

Similarly, within the eight qualifying studies, not all model implementations reported could be used: several ensembles, secondary reporting of other studies’ results, and implementations not evaluated on each of the 10 items were not included in the data. All primary results referenced in papers reporting secondary results were also found in the original search. The results of this search were also used to add to the benchmarking chart done by \citeauthor{kiela_plotting_2023} seen in Figure \ref{fig:benchmarks}.


While we necessarily call attention to \citeauthor{sivakumar_future_2026} and \citeauthor{todorov_evaluating_2025} in Section \ref{sec:removal_siva} in order to conduct our study, these researchers were not alone in errors. Without naming names or unduly calling out venues or researchers encountered or missed in our systematic review, we found that far more than 90\% of publications utilizing the ``ASAP-SAS'' dataset to demonstrate their experiments did not meet the criteria, and those excluded were generally poor quality research: they found ways to justify not using QWK (including creating bespoke metrics), relying on accuracy (which is easily inflated and misleading in this imbalanced dataset), reporting only custom aggregations (without access to the underlying or specific data leading to their construction), or cherry‑picked subsets of items (especially, those subsets that excluded the hardest for LLMs to grade). These include the sizable number who claimed some form of SOTA results which were verifiably not SOTA at the time of publication. Nevertheless, we applaud both \citeauthor{sivakumar_future_2026} and \citeauthor{todorov_evaluating_2025}, because despite their errors, they provided a full reporting, including what might be considered null results, and offered their prompting techniques for scrutiny. After reviewing dozens of poor papers whose creativity is greatest in finding ways to avoid generalizable science, we recommend that the community establish some baseline requirements for using datasets as benchmarks (including reporting the metrics intended to be used, even if it might make research findings look weaker). We choose not to cite the other studies that did not meet the minimal criteria.

\subsubsection{Removal of Observations}\label{sec:removal_siva}
We needed to exclude all results from item number 3 from the dataset of \citeauthor{sivakumar_future_2026}, as the results, which exactly match the range of values one would get from stratified random sampling of the item responses, almost certainly represent a mismatch of scores between the samples tested and the true labels. Indeed, using the prompts found in the appendix, we cannot reproduce their results for this item on GPT-4 or GPT-4o. 

We similarly exclude the LoRA adaptations and baseline BERT values from \citeauthor{todorov_evaluating_2025} from Table 4 of that paper, as they represent values too close to random chance (using methods that seem technologically misapplied to these data). Nevertheless we applaud them reporting the null result from their exploration and praise them even more for releasing their code and experiments on github.\footnote{\url{https://github.com/atodorov284/automatic_sas_with_llms}} 

While we needed to remove observations that followed the same distributions as stratified random sampling, we are happy to cite both of these papers for their willingness to produce replicable science, which, embarrassingly for the community of researchers interested in the topic, and appears to be rare and with little to show for not doing so. Indeed, the hypothesis of the proliferation of low quality research in this area were part of the original impetus for the present study, but we had underestimated the extent of these challenges. 

We hope that this unobtrusive appendical rebuke is sufficient for improving the research and reporting of one of the most important topics of educational research.

\subsection{Final Dataset}
The final metadataset from the systemic review can be found in Table \ref{tab:metadataset}.

\begin{table*}
\centering
\caption{Included Studies and Properties of metadataset}
\label{tab:metadataset}
\resizebox{\ifdim\width>\linewidth\linewidth\else\width\fi}{!}{
\begin{tabular}{lcccccccc}
\toprule
  \textbf{Characteristic} & Jiang \& Bosch & Kim et al. & Ormerod & Ormerod \& Kwako & Sivkumar et al. & Surya et al. & Todorov et al. & Zeng et al. \\
 Year & 2024 & 2025 & 2022 & 2024 & 2026 & 2019 & 2025 & 2023 \\
 N & 60 & 90 & 130 & 80 & 360 & 10 & 40 & 120 \\

\midrule
qwk & 0.63 (0.52, 0.70) & 0.72 (0.65, 0.80) & 0.76 (0.70, 0.82) & 0.61 (0.34, 0.75) & 0.45 (0.30, 0.58) & 0.72 (0.66, 0.79) & 0.74 (0.68, 0.79) & 0.69 (0.61, 0.73)\\
token &  &  &  &  &  &  &  & \\
\hspace{1em}bpe & 60 (100\%) & 70 (78\%) & 20 (15\%) & 80 (100\%) & 300 (83\%) & 0 (0\%) & 0 (0\%) & 0 (0\%)\\
\hspace{1em}uni & 0 (0\%) & 0 (0\%) & 30 (23\%) & 0 (0\%) & 60 (17\%) & 0 (0\%) & 0 (0\%) & 0 (0\%)\\
\hspace{1em}wp & 0 (0\%) & 20 (22\%) & 80 (62\%) & 0 (0\%) & 0 (0\%) & 10 (100\%) & 40 (100\%) & 120 (100\%)\\
\addlinespace
model &  &  &  &  &  &  &  & \\
\hspace{1em}ALBERT & 0 (0\%) & 0 (0\%) & 10 (7.7\%) & 0 (0\%) & 0 (0\%) & 0 (0\%) & 0 (0\%) & 0 (0\%)\\
\hspace{1em}BERT & 0 (0\%) & 10 (11\%) & 20 (15\%) & 0 (0\%) & 0 (0\%) & 10 (100\%) & 40 (100\%) & 120 (100\%)\\
\hspace{1em}CBERT & 0 (0\%) & 0 (0\%) & 10 (7.7\%) & 0 (0\%) & 0 (0\%) & 0 (0\%) & 0 (0\%) & 0 (0\%)\\
\hspace{1em}DistilBERT & 0 (0\%) & 0 (0\%) & 10 (7.7\%) & 0 (0\%) & 0 (0\%) & 0 (0\%) & 0 (0\%) & 0 (0\%)\\
\hspace{1em}DeBERTa & 0 (0\%) & 10 (11\%) & 10 (7.7\%) & 0 (0\%) & 0 (0\%) & 0 (0\%) & 0 (0\%) & 0 (0\%)\\
\addlinespace
\hspace{1em}Electra & 0 (0\%) & 0 (0\%) & 30 (23\%) & 0 (0\%) & 0 (0\%) & 0 (0\%) & 0 (0\%) & 0 (0\%)\\
\hspace{1em}Gemma & 0 (0\%) & 0 (0\%) & 0 (0\%) & 20 (25\%) & 0 (0\%) & 0 (0\%) & 0 (0\%) & 0 (0\%)\\
\hspace{1em}GPT 3.5 & 0 (0\%) & 0 (0\%) & 0 (0\%) & 0 (0\%) & 60 (17\%) & 0 (0\%) & 0 (0\%) & 0 (0\%)\\
\hspace{1em}GPT 4o & 0 (0\%) & 0 (0\%) & 0 (0\%) & 0 (0\%) & 60 (17\%) & 0 (0\%) & 0 (0\%) & 0 (0\%)\\
\hspace{1em}GPT4 & 60 (100\%) & 0 (0\%) & 0 (0\%) & 0 (0\%) & 60 (17\%) & 0 (0\%) & 0 (0\%) & 0 (0\%)\\
\addlinespace
\hspace{1em}GPT4.1 mini & 0 (0\%) & 20 (22\%) & 0 (0\%) & 0 (0\%) & 0 (0\%) & 0 (0\%) & 0 (0\%) & 0 (0\%)\\
\hspace{1em}LlaMA & 0 (0\%) & 50 (56\%) & 0 (0\%) & 20 (25\%) & 0 (0\%) & 0 (0\%) & 0 (0\%) & 0 (0\%)\\
\hspace{1em}MobileBERT & 0 (0\%) & 0 (0\%) & 10 (7.7\%) & 0 (0\%) & 0 (0\%) & 0 (0\%) & 0 (0\%) & 0 (0\%)\\
\hspace{1em}Mistral & 0 (0\%) & 0 (0\%) & 0 (0\%) & 20 (25\%) & 0 (0\%) & 0 (0\%) & 0 (0\%) & 0 (0\%)\\
\hspace{1em}Mistral large & 0 (0\%) & 0 (0\%) & 0 (0\%) & 0 (0\%) & 60 (17\%) & 0 (0\%) & 0 (0\%) & 0 (0\%)\\
\addlinespace
\hspace{1em}Mixtral 8x22b & 0 (0\%) & 0 (0\%) & 0 (0\%) & 0 (0\%) & 60 (17\%) & 0 (0\%) & 0 (0\%) & 0 (0\%)\\
\hspace{1em}Palm 2 & 0 (0\%) & 0 (0\%) & 0 (0\%) & 0 (0\%) & 60 (17\%) & 0 (0\%) & 0 (0\%) & 0 (0\%)\\
\hspace{1em}Phi & 0 (0\%) & 0 (0\%) & 0 (0\%) & 20 (25\%) & 0 (0\%) & 0 (0\%) & 0 (0\%) & 0 (0\%)\\
\hspace{1em}RoBERTa & 0 (0\%) & 0 (0\%) & 20 (15\%) & 0 (0\%) & 0 (0\%) & 0 (0\%) & 0 (0\%) & 0 (0\%)\\
\hspace{1em}XLNet & 0 (0\%) & 0 (0\%) & 10 (7.7\%) & 0 (0\%) & 0 (0\%) & 0 (0\%) & 0 (0\%) & 0 (0\%)\\
training &  &  &  &  &  &  &  & \\
\hspace{1em}fine-tune & 0 (0\%) & 20 (22\%) & 130 (100\%) & 0 (0\%) & 0 (0\%) & 10 (100\%) & 40 (100\%) & 120 (100\%)\\
\hspace{1em}LORA+prompt & 0 (0\%) & 40 (44\%) & 0 (0\%) & 40 (50\%) & 0 (0\%) & 0 (0\%) & 0 (0\%) & 0 (0\%)\\
\hspace{1em}multi-prompt & 10 (17\%) & 20 (22\%) & 0 (0\%) & 0 (0\%) & 0 (0\%) & 0 (0\%) & 0 (0\%) & 0 (0\%)\\
\hspace{1em}prompt engineer & 50 (83\%) & 10 (11\%) & 0 (0\%) & 40 (50\%) &  360 (100\%)  & 0 (0\%) & 0 (0\%) & 0 (0\%)\\
finetune & 0 (0\%) & 60 (67\%) & 130 (100\%) & 40 (50\%) & 0 (0\%) & 10 (100\%) & 40 (100\%) & 120 (100\%)\\
log B param. & 3.25 (3.25, 3.25) & 0.90 (0.90, 0.90) & -0.93 (-1.17, -0.65) & 0.88 (0.72, 0.92) & 2.27 (2.15, 2.53) & -0.96 (-0.96, -0.96) & -0.96 (-0.96, -0.96) & -0.96 (-0.96, -0.96)\\
\bottomrule
\multicolumn{9}{l}{\rule{0pt}{1em}\textsuperscript{1} Median (Q1, Q3); n (\%)}\\
\end{tabular}

}

\end{table*}

\section{Unintended Biases}\label{apx:gpt}
Below is the full prompt and responses used to generate the examples in Table \ref{tab:gpt}. The task and student response are from the Smarter Balanced student writing examples for third grade.  The same prompt was used twice, replacing *** with “White” and “Black”, respectively, on May 1, 2024 using the default ChatGPT engine interface. The table is not meant to represent cutting edge technologies: only to demonstrate how easy it is to elicit biases through obvious language usage.

\texttt{You are an accurate and helpful writing bot. In this scenario, you will provide a numeric rating and feedback to diagnose the English writing Conventions for a written response.
\\
\\
<<STUDENT TASK>> Third grade students were asked to respond to the following task:
\\
\\
"You will write a story about an astronaut that tells about an adventure that happens to a character who is an astronaut traveling in space. Your story will be read by parents, teachers, and other students.
\\
\\
Writers often do research to add realistic detail to the setting, characters, and plot in their stories. When writing your story, find ways to use information and details from the sources to improve your story about an astronaut adventure. Make sure you develop your characters, the setting, and the plot. Use details, dialogue, and description where appropriate.
\\
\\
REMEMBER: A well-written story has a clear plot and a clear sequence of events, is well-organized and has a point of view, uses details from the sources to improve the story, uses clear language, and follows rules of writing (spelling, capitalization, punctuation, and grammar usage)."
\\
\\
<<STUDENT RESPONSE>> Here is the written response you will provide a rating for:
““
“5”…”4”…”3”…”2” “1”… “BLASToff!”
“I’m very exited to go in space” exclaimed Even
“Don’t get to exited” Joe said calmly
“Why” anctiqued Even  “Oh wait it’s because that Even explained
youre body changes and you need to exersiez”
“Yes” Joe said. When they were in space
“Joe what if we crash on an alian’s planet” Even asked “then…RUN” Joe screamed.
“can you keep it down?!” Hannah said angerly ”Yea” replied Emma  “I hope I don’t get the condition bird legs” Even said nervously  “Oh you won’t we are not going to be at the moon for long we are just resting on the moon and taking some small rocks and bringing them back” Joe explained.
“Also bird legs is not that bad” said Emma “Seriously do you even know what the condition bird legs is” “yes” Emma said anoyed “then what is it” Even said
Emma replied “bird legs is a condition when blood flows to youre legs making then skinnier” “Not that I think about it it’s still bad  but not as bad as other conditions” Even said. “We are crashing” the pilot yelled. Everyone “AHHHHHHH” when they crashed “Where are we” Emma whispered (and they do have their space suits on) “I don’t know” Even whispered back “AHHH” Hannah screamed “HANNAH” Even Emma and Joe screamed “OH NO We are on an alein’s planet” Even said loudly “Well be quiet we have to whisper” Emma whispered. “Now to save Hannah” after they saved Hannah now let’s go home” Even said “I finnaly fixed it” Even replied happy
When they got home Evens mind I want to go there again.
””
\\ 
\\
<<YOUR INSTRUCTIONS>>
This written response was created by a *** student in Mrs. Jones’ class. Provide a numeric rating from 0 to 2 to classify the quality of developmentally appropriate English written Conventions of this written response and provide feedback for your numeric rating. 
\\ 
\\
<<YOUR NUMERIC RATING FOR CONVENTIONS>>
}

\section{Main Body Estimations}
In this section we provide additional details and results regarding the models in the main body of the paper. Additional, alternative estimations can be found in Section \ref{apx:addlests}, which fit various other models, including with and without data flagged as incorrect.

\subsubsection{Anova Table}
Table \ref{tab:anova_main} has the results from analysis of variance comparisons.
\begin{table*}
\centering
\resizebox{\ifdim\width>\linewidth\linewidth\else\width\fi}{!}{
\begin{tabular}{lrrrrrrrr}
\toprule
term & npar & AIC & BIC & log($\mathcal{L}$) & -2log($\mathcal{L}$) & statistic & df & p.value\\
\midrule
(1) & 12 & -359.95 & -302.46 & 191.98 & -383.95 & NA & NA & NA\\
(2) & 13 & -418.17 & -355.88 & 222.09 & -444.17 & 60.22 & 1 & 0.00\\
(3) & 14 & -711.65 & -644.58 & 369.83 & -739.65 & 295.48 & 1 & 0.00\\
(4) & 20 & -974.69 & -878.87 & 507.35 & -1014.69 & 275.04 & 6 & 0.00\\
(5) & 23 & -969.24 & -859.04 & 507.62 & -1015.24 & 0.55 & 3 & 0.91\\
\addlinespace
(6) & 175 & -845.64 & -7.17 & 597.82 & -1195.64 & 180.40 & 152 & 0.06\\
\bottomrule
\end{tabular}}
\caption{Anova results for main body models.}
\label{tab:anova_main}
\end{table*}

\subsection{Full Bayesian Parameter Estimations}
In this section we report completely on the Bayesian estimation of the item-interaction random effects model described in Equation \ref{eq:m6} and summarized in the right column of Table \ref{tab:metaregsummary}.

\subsubsection{Priors and computation}\label{apx:bayes_priors}
We used noninformative and weakly informative regularizing priors (brms defaults) for fixed effects and variance components, which shrink extreme values toward plausible ranges while allowing the data to dominate when information is strong. Inference used Hamiltonian Monte Carlo (CmdStan) with 5 chains, 4000 iterations per chain (1000 warmup), thinning 5, \texttt{adapt\_delta}=0.95, and \texttt{max\_treedepth}=15. Convergence was assessed via ($\hat{R}\approx 1.00$) and effective sample sizes for both bulk and tails. Priors for all independent variables were flat,  LKJ correlation matrix densities (Cholesky factor $\eta=1$) were used for correlational relationships, and all random variables used a weakly informative Student's t distribution: $y \sim t(\nu=3, \mu=0, \sigma=2.5)$.

\begin{table*}

\centering
\caption{Anova results for Appendical Nine models fit on selected data.}
\resizebox{\ifdim\width>\linewidth\linewidth\else\width\fi}{!}{

\begin{tabular}{lrrrrrrr}
\toprule
model& npar & AIC & BIC & logLik & stat.& df & p.val\\
\midrule
m1 & 11 & -495.89 & -449.75 & 258.95 &   &   &  \\
m2 & 12 & -494.00 & -443.67 & 259.00 & 0.11 & 1 & 0.74\\
m3 & 12 & -493.96 & -443.63 & 258.98 & 0.00 & 0 &  \\
m4 & 13 & -492.07 & -437.54 & 259.04 & 0.11 & 1 & 0.74\\
m5 & 15 & -488.15 & -425.23 & 259.08 & 0.08 & 2 & 0.96\\
m6 & 20 & -506.12 & -422.23 & 273.06 & 27.97 & 5 & 0.00\\
m7 & 24 & -567.30 & -466.63 & 307.65 & 69.18 & 4 & 0.00\\
m8 & 25 & -565.23 & -460.37 & 307.61 & 0.00 & 1 & 1.00\\
m9 & 27 & -562.01 & -448.77 & 308.01 & 0.79 & 2 & 0.67\\
\bottomrule

\end{tabular}

}

\end{table*}

\section{Additional Models}
\subsection{Additional Estimations}\label{apx:addlests}
In the subsequent tables, we display the estimates of nine additional sequentially complex models and note the stability of the estimates, consistent with the models presented in the main body.

In section \ref{sec:removal_siva}, we discuss weaknesses of the studies included in the main body. In this section, we estimate various permutations of models after removing data sourced from \citeauthor{sivakumar_future_2026} and \citeauthor{todorov_evaluating_2025} from the datasets. 

Let $y_{ij}$ denote effect-size $\mathrm{QWK}$ for model implementation i on item set j. We fit nine mixed-effects models which follow the general format $y_{ij} = \beta_{0} + \beta_{1}\,\mathrm{Reading}_{j} + \beta_{2}\,\mathrm{Decoder}_{i} + f(\mathrm{VocabSize}_{i}) + u_{i} + v_{j} + \varepsilon_{ij}$. 
Four more complex models found in the rightmost columns of Table \ref{tab:9mods} follow a more complex format, found in Section \ref{apx:gpt}, 
$y_{ij} = \beta_{0} + \beta_{1}\,\mathrm{Reading}_{j} + \beta_{2}\,\mathrm{Decoder}_{i} + f(\mathrm{VocabSize}_{i}) + u_{i} + v_{j} + \varepsilon_{ij}$,
with random intercepts $u_{i}$ (implementation) and $v_{j}$ (item set) to absorb unobserved heterogeneity, and $\varepsilon_{ij}$ as residual error.  Table \ref{tab:9summ} provides a summary of four models representing these distinct features and show the stability of estimates. 
The fixed effects encode three testable hypotheses: (1) meaning-intensive reading items depress performance $(\beta_{1} < 0)$; (2) decoder-only GPT-style models underperform encoder-based implementations $(\beta_{2} < 0)$; (3) tokenizer vocabulary size shows diminishing returns, captured via a concave $f(\cdot)$, e.g., $f(x) = \beta_{3}\log x + \beta_{4}(\log x)^{2}$ with $\beta_{4} < 0$. We evaluate robustness across prompt-engineered and fine-tuned settings and compare against human rater agreement. All these statistical tests were run in \texttt{R} \cite{r_core_team_r_nodate} using the \texttt{lme4} package \cite{bates_fitting_2015}. Model estimations used the BOBYQA algorithm for optimization \cite{powell_bobyqa_2009} and all converged in $\le 1e7$ max iterations. The equations listed are represented in \texttt{R} syntax in their respective order below.

\begin{table}
\centering\centering
\caption{Quick Summary of Four Models from Robustness Checks}
\label{tab:9summ}

\resizebox{\ifdim\width>\linewidth\linewidth\else\width\fi}{!}{
\begin{tabular}[t]{lcccc}
\toprule
  & m1 & m6 & m7 & m9\\
\midrule
(Intercept) & 0.69** & 0.73*** &  & \\
 & [0.25, 0.89] & [0.43, 0.88] &  & \\
$\operatorname{QWK}_\text{hum}$ & 0.06 & 0.06 & -0.06 & -0.06\\
 & [-0.17, 0.29] & [-0.17, 0.29] & [-0.21, 0.08] & [-0.20, 0.09]\\
read & -0.20* & -0.21** & -0.24*** & -0.24***\\
 & [-0.35, -0.05] & [-0.35, -0.05] & [-0.33, -0.15] & [-0.37, -0.11]\\
$\text{|tok. voc.|}$  & 0.27* & 0.17* & 0.18** & 0.18**\\
 & [0.07, 0.45] & [0.03, 0.30] & [0.05, 0.31] & [0.05, 0.31]\\

$\text{|tok. voc.|}^2$ & -0.12** & -0.07** & -0.08** & -0.08**\\
 & [-0.20, -0.03] & [-0.12, -0.02] & [-0.13, -0.03] & [-0.13, -0.03]\\
logsize &  & -0.01+ & -0.01* & -0.01*\\
 &  & [-0.03, 0.00] & [-0.03, -0.00] & [-0.03, -0.00]\\
 gpt & -0.23 & -0.14 & -0.14 & -0.28*\\
 & [-0.63, 0.25] & [-0.33, 0.06] & [-0.33, 0.06] & [-0.48, -0.05]\\
read × gpt &  & 0.01 &  & -0.00\\
 &  & [-0.04, 0.05] &  & [-0.08, 0.07]\\
$\text{tok}_\text{bpe}$ &  &  & 0.81*** & 0.83***\\
 &  &  & [0.66, 0.90] & [0.69, 0.91]\\
$\text{tok}_\text{uni}$ &  & -0.02 & 0.80*** & 0.83***\\
 &  & [-0.10, 0.07] & [0.65, 0.89] & [0.69, 0.91]\\
$\text{tok}_\text{wp}$ &  & -0.00 & 0.81*** & 0.83***\\
 &  & [-0.09, 0.08] & [0.65, 0.90] & [0.69, 0.91]\\
 \midrule
\textbf{\textit{Rand. Effect}}  &  &  &  &  \\
LLM & Y & Y & Y & Y\\
Study & Y & Y & Y & Y\\
Training & Y & Y & Y & Y\\
Item & Y & Y & Y & Y\\
\textbf{\textit{Rand. Slope}}  &  &  &  & \\
Tokenizer &  &  & Item & Item \\ 
GPT LLMs &  &  & Train & Train \\
GPT LLMs &  &  &  & Study \\
logsize &  &  &  & Train \\

\midrule
R2 Cond. & 0.846 &  0.865&  0.869& 0.912\\
AIC & -477.2 & -458.5 & -524.9 & -514.9\\
\bottomrule
\end{tabular}

}

\vspace{0.25mm}
+ p < 0.1, * p < 0.05, ** p < 0.01, *** p < 0.001. 
These summarize findings of the complete Table \ref{tab:9mods}.

\end{table}

\begin{table*}
\centering\centering
\caption{Full Estimate Table for nine more appendical models under variable permutations.} \label{tab:9mods}

\resizebox{\ifdim\width>\linewidth\linewidth\else\width\fi}{!}{
\begin{tabular}[t]{lccccccccc}
\toprule
  & m1 & m2 & m3 & m4 & m5 & m6 & m7 & m8 & m9\\
\midrule
(Intercept) & 0.69** & 0.69** & 0.69** & 0.69** &  & 0.73*** &  &  & \\
 & [0.25, 0.89] & [0.25, 0.89] & [0.25, 0.89] & [0.25, 0.89] &  & [0.43, 0.88] &  &  & \\
$\operatorname{QWK}_\text{hum}$ & 0.06 & 0.06 & 0.06 & 0.06 & 0.06 & 0.06 & -0.06 & -0.06 & -0.06\\
 & [-0.17, 0.29] & [-0.17, 0.29] & [-0.17, 0.29] & [-0.17, 0.29] & [-0.17, 0.29] & [-0.17, 0.29] & [-0.21, 0.08] & [-0.20, 0.09] & [-0.20, 0.09]\\
read & -0.20* & -0.20* & -0.21** & -0.21** & -0.21** & -0.21** & -0.24*** & -0.24*** & -0.24***\\
 & [-0.35, -0.05] & [-0.35, -0.05] & [-0.35, -0.05] & [-0.35, -0.05] & [-0.35, -0.05] & [-0.35, -0.05] & [-0.33, -0.15] & [-0.37, -0.10] & [-0.37, -0.11]\\
$\text{|tok. voc.|}$  & 0.27* & 0.27* & 0.27* & 0.27* & 0.27* & 0.17* & 0.18** & 0.18** & 0.18**\\
 & [0.07, 0.45] & [0.06, 0.45] & [0.07, 0.45] & [0.06, 0.45] & [0.06, 0.46] & [0.03, 0.30] & [0.05, 0.31] & [0.05, 0.31] & [0.05, 0.31]\\

$\text{|tok. voc.|}^2$ & -0.12** & -0.12** & -0.12** & -0.12** & -0.12** & -0.07** & -0.08** & -0.08** & -0.08**\\
 & [-0.20, -0.03] & [-0.20, -0.03] & [-0.20, -0.03] & [-0.20, -0.03] & [-0.20, -0.03] & [-0.12, -0.02] & [-0.13, -0.03] & [-0.13, -0.03] & [-0.13, -0.03]\\
logsize &  & 0.00 &  & 0.00 & 0.00 & -0.01+ & -0.01* & -0.01* & -0.01*\\
 &  & [-0.01, 0.01] &  & [-0.01, 0.01] & [-0.01, 0.01] & [-0.03, 0.00] & [-0.03, -0.00] & [-0.03, -0.00] & [-0.03, -0.00]\\
 gpt & -0.23 & -0.24 & -0.24 & -0.24 & -0.24 & -0.14 & -0.14 & -0.14 & -0.28*\\
 & [-0.63, 0.25] & [-0.63, 0.25] & [-0.63, 0.25] & [-0.63, 0.25] & [-0.65, 0.27] & [-0.33, 0.06] & [-0.33, 0.06] & [-0.34, 0.07] & [-0.48, -0.05]\\
read × gpt &  &  & 0.01 & 0.01 & 0.01 & 0.01 &  & -0.00 & -0.00\\
 &  &  & [-0.04, 0.05] & [-0.04, 0.05] & [-0.04, 0.05] & [-0.04, 0.05] &  & [-0.08, 0.07] & [-0.08, 0.07]\\
$\text{tok}_\text{bpe}$ &  &  &  &  & 0.69** &  & 0.81*** & 0.81*** & 0.83***\\
 &  &  &  &  & [0.23, 0.90] &  & [0.66, 0.90] & [0.65, 0.90] & [0.69, 0.91]\\
$\text{tok}_\text{uni}$ &  &  &  &  & 0.68** & -0.02 & 0.80*** & 0.80*** & 0.83***\\
 &  &  &  &  & [0.22, 0.89] & [-0.10, 0.07] & [0.65, 0.89] & [0.64, 0.89] & [0.69, 0.91]\\
$\text{tok}_\text{wp}$ &  &  &  &  & 0.69** & -0.00 & 0.81*** & 0.80*** & 0.83***\\
 &  &  &  &  & [0.25, 0.89] & [-0.09, 0.08] & [0.65, 0.90] & [0.65, 0.90] & [0.69, 0.91]\\
$\sigma_{\text{LLM}}$ & 0.07 & 0.07 & 0.07 & 0.07 & 0.07 & 0.02 & 0.03 & 0.03 & 0.03\\
$\sigma_{\text{study}}$ & 0.08 & 0.07 & 0.08 & 0.07 & 0.08 & 0.17 & 0.18 & 0.18 & 0.13\\
$\sigma_{\text{train}}$& 0.21 & 0.21 & 0.21 & 0.21 & 0.21 & 0.12 & 0.12 & 0.12 & 0.13\\
$\sigma_{\text{item}}(\text{Intercept}) $ & 0.10 & 0.10 & 0.10 & 0.10 & 0.10 & 0.10 &  &  & \\
$\sigma_{\text{item}}(\text{tok}_\text{bpe})$&  &  &  &  &  &  & 0.07 & 0.07 & 0.07\\
$\sigma_{\text{item}}(\text{tok}_\text{uni})$&  &  &  &  &  &  & 0.12 & 0.11 & 0.11\\
$\sigma_{\text{item}}(\text{tok}_\text{wp})$ &  &  &  &  &  &  & 0.15 & 0.15 & 0.15\\
\addlinespace
$\sigma_{\text{study}}$(gpt) &  &  &  &  &  &  &  &  & 0.09\\
$\sigma_{\text{train}}$(gpt) &  &  &  &  &  & 0.16 & 0.17 & 0.17 & 0.19\\
$\sigma_{\text{train}}$(logsize)  &  &  &  &  &  & 0.01 & 0.01 & 0.01 & 0.01\\
\addlinespace
$\text{Cor}_\text{study}$(Intcpt$\sim$ gpt) &  &  &  &  &  &  &  &  & 0.76\\
$\text{Cor}_\text{train}$(Intcpt$\sim$ gpt)  &  &  &  &  &  & 0.76 & 0.76 & 0.76 & 0.76\\
$\text{Cor}_\text{train}$(Intcpt$\sim$logsize) &  &  &  &  &  & -0.76 & -0.76 & -0.76 & -0.76\\
$\text{Cor}_\text{train}$(gpt$\sim$logsize) &  &  &  &  &  & -0.76 & -0.76 & -0.76 & -0.76\\
$\text{Cor}_\text{item}$($\text{tok}_\text{bpe}\sim\text{tok}_\text{uni})$ &  &  &  &  &  &  & 0.76 & 0.76 & 0.76\\
$\text{Cor}_\text{item}$($\text{tok}_\text{bpe}\sim\text{tok}_\text{wp})$ &  &  &  &  &  &  & 0.68 & 0.67 & 0.67\\
$\text{Cor}_\text{item}$($\text{tok}_\text{uni}\sim\text{tok}_\text{wp})$ &  &  &  &  &  &  & 0.65 & 0.67 & 0.67\\
$\sigma_\epsilon$ & 0.13 & 0.13 & 0.13 & 0.13 & 0.13 & 0.13 & 0.12 & 0.12 & 0.12\\
\midrule
Num.Obs. & 490 & 490 & 490 & 490 & 490 & 490 & 490 & 490 & 490\\
R2 Marg. & 0.232 & 0.234 & 0.232 & 0.234 & 0.235 & 0.259& 0.266& 0.266& 0.346\\
R2 Cond. & 0.846 & 0.845 & 0.846 & 0.845 & 0.846 &  0.865&  0.869&  0.900& 0.912\\
AIC & -477.2 & -465.8 & -469.6 & -458.3 & -447.4 & -458.5 & -524.9 & -518.1 & -514.9\\
BIC & -431.0 & -415.5 & -419.3 & -403.8 & -384.5 & -374.7 & -424.2 & -413.3 & -401.7\\
RMSE & 0.13 & 0.13 & 0.13 & 0.13 & 0.13 & 0.13 & 0.11 & 0.11 & 0.11\\
\bottomrule
\multicolumn{10}{l}{\rule{0pt}{1em}+ p < 0.1, * p < 0.05, ** p < 0.01, *** p < 0.001}\\
\end{tabular}

}
\vspace{0.25cm}

\end{table*}

\end{document}